\documentclass[10pt,twocolumn,letterpaper]{article}

\usepackage{wacv}
\usepackage{times}
\usepackage{epsfig}
\usepackage{graphicx}
\usepackage{amsmath}
\usepackage{amssymb}

\usepackage{dsfont}

\usepackage[ruled,vlined]{algorithm2e}
\usepackage{enumitem}

\usepackage{stmaryrd}
\usepackage{textcomp}

\usepackage{multirow}
\usepackage{multicol}
\usepackage{makecell}
\usepackage{float}

\DeclareMathOperator*{\argmax}{arg\,max}

\def\wacvPaperID{981} 

\wacvfinalcopy 

\ifwacvfinal
\fi

\ifwacvfinal
\usepackage[breaklinks=true,bookmarks=false]{hyperref}
\else
\usepackage[pagebackref=true,breaklinks=true,colorlinks,bookmarks=false]{hyperref}
\fi

\begin{document}

\title{Unveiling Real-Life Effects of Online Photo Sharing}

\author{Van-Khoa Nguyen, Adrian Popescu, Jérôme Deshayes-Chossart\\
Université Paris-Saclay, CEA, List, F-91120, Palaiseau, France\\
{\tt\small khoa.v18nguyen@gmail.com,\{adrian.popescu,jerome.deshayes-chossart\}@cea.fr}
}

\maketitle
\thispagestyle{empty}

\begin{abstract}
Social networks give free access to their services in exchange for the right to exploit their users' data.
Data sharing is done in an initial context which is chosen by the users.
However, data are used by social networks and third parties in different contexts which are often not transparent.
In order to unveil such usages, we propose an approach which focuses on the effects of data sharing in impactful real-life situations.
Focus is put on visual content because of its strong influence in shaping online user profiles.
The approach relies on three components: (1) a set of visual objects with associated situation impact ratings obtained by crowdsourcing, (2) a corresponding set of object detectors for mining users' photos and (3) a ground truth dataset made of 500 visual user profiles which are manually rated per situation. 
These components are combined in $LERVUP$, a method which learns to rate visual user profiles in each situation.
$LERVUP$ exploits a new image descriptor which aggregates object ratings and object detections at user level and an attention mechanism which boosts highly-rated objects to prevent them from being overwhelmed by low-rated ones.
Performance is evaluated per situation by measuring the correlation between the automatic ranking of profile ratings and a manual ground truth. 
Results indicate that $LERVUP$ is effective since a strong correlation of the two rankings is obtained. 
A practical implementation of the approach in a mobile app which raises user awareness about shared data usage is also discussed.

\end{abstract}

\vspace{-5mm}
\section{Introduction}
The ubiquitous use of online social networks (OSNs) shows that their services are appealing to users.
Most OSNs implement a business model in which access is free in exchange for user data monetization~\cite{curran2011advertising}.
Intrusiveness is likely to grow with the wide usage of AI techniques to infer actionable information from users' data. 
Automatic inferences happen in the back-end of OSNs or of associated third parties and are not transparent for users.
Data can be exploited in contexts unforeseen when sharing them initially.
The main objective of our work is to improve user awareness about data processing through feedback contextualization. 
To do this, we introduce a plausible decision-making system which combines machine learning and domain knowledge. 

User awareness is increased by linking the sharing process to impactful situations such as searching for a job, an accommodation, or a bank credit.
Photos are in focus because they constitute a large part of shared data and contribute strongly to shaping user profiles~\cite{ahern2007over}.
The main technical contribution is a method that rates visual user profiles and individual photos in a given situation by exploiting situation models, visual detectors and a dedicated photographic profiles dataset.
The proposed method, named $LERVUP$ from \textbf{LE}arning to \textbf{R}ate \textbf{V}isual \textbf{U}ser \textbf{P}rofiles, learns a ranking of user profiles which attemps to reproduce human profiles ranking.
$LERVUP$ exploits a new descriptor which combines object impact ratings and object detections in a compact form.
The contributions of objects with high ratings are boosted in order to mimic the way humans assess photographic content. 
We compare manual and automatic rankings of user profile ratings and obtain a positive correlation between them.
This result holds promise to help users better understand the effects of online data sharing and, ultimately, to better control their data.

\section{Related work}
\label{sec:sota}
The main promise of OSNs is to connect people and allow them to exchange information within affinity-based networks.
Participation in OSNs can have positive or negative effects, depending on the way the shared information is interpreted in different contexts~\cite{burkesocial}.
This interpretation process is influenced by human and/or technical biases.
Human biases fall in two main categories that are often studied in relation to demographic factors such as gender or ethnicity~\cite{williams2014double}.
Implicit biases~\cite{greenwald1995implicit} might influence one's decisions without that person being conscious of them.
Explicit biases~\cite{clarke2018explicit} are assumed and used intentionally.
Technical imperfections lead to the occurrence of algorithmic biases.
The inherently partial mapping of complex real-life processes into computer systems~\cite{friedman1996bias} is a first source of bias. 
For instance, the introduction of deep architectures brought important progress in classification~\cite{cirecsan2012multi,krizhevsky2017imagenet}, but accuracy is still affected by internal representation limitations~\cite{nguyen2015deep}.
A second bias is due to the variable accuracy of predictions due to the inherent difficulty of visual objects~\cite{cai2016unified} and/or the availability of skewed~\cite{tommasi2017deeper} and imbalanced data~\cite{huang2016learning}.
While affected by biases, approaches like ours are needed to make AI-powered decision-making more transparent to users.
Notably, impact and profile ratings are potentially biased toward the opinions of the persons involved in the experiments.
Object detection is biased because the detection dataset is incomplete and detectors are imperfect.

Below, we illustrate contexts in which users' lives are influenced by their online activity. 
The authors of~\cite{acquisti2020} and ~\cite{manant2019} create fictitious Facebook profiles in which they vary only one type of personal data to assess its influence during a job search.
No significant discrimination due to family structure and sexual orientation is found in~\cite{acquisti2020}, while a negative effect is elicited for radical religious stance.
The user's supposed origin has a significant effect on the number of replies a person gets to a job application~\cite{manant2019}.
The chances to obtain short-term accommodation online are influenced by the assumed racial origin~\cite{edelman2017racial}.
Rather accurate creditworthiness is automatically obtained in~\cite{credit_scoring} based on one's interests and the analysis of the list of friends.
These studies focus on sensitive signals and contributed to a degree of public awareness about their effects. 
Our objective is to include weaker signals for which there is little or no awareness. Such signals seem innocuous in the initial sharing context, but their interpretation might change in other situations, and users should be informed about such changes. 

The prediction of user traits from shared data received much attention in the last decade.
The threats induced by geolocation mining were studied in~\cite{DBLP:journals/ijsc/FriedlandC11}.
In~\cite{Kosinski5802}, Facebook likes were exploited to predict sexual orientation, political opinions, race and personality traits. 
A hierarchical organization of privacy aspects and methods that predict privacy traits were proposed in~\cite{petkos2015}.
The authors of~\cite{diazferreyra:hal-01677149} implemented an instructional awareness system which provides feedback about content whose publishing might be harmful to the users.
These works explore interesting aspects of privacy but do not provide a systematic way to map predictions to real-life situations and do not focus on visual data.

The understanding of the effects of visual content sharing was pioneered by~\cite{ahern2007over}, with the introduction of disclosure dimensions such as security, identity and convenience.
The study concludes that user feedback should provide warnings to prevent mistakes, inform about the effects of data aggregation and estimate the appropriate audience.
The authors of~\cite{zerr2012} used hand-crafted visual features to predict the privacy status of an image with encouraging results but far from practical usability. 
Transfer learning from generalist deep models as a way to improve privacy prediction was proposed in~\cite{spyromitrosxioufis2016}.
An important step forward was made in~\cite{orekondy2017}, with the creation of a taxonomy of privacy-related attributes and of a dataset dedicated to privacy prediction. 
Interestingly, the resulting model provided more consistent predictions compared to users' judgments, indicating that users might fail to follow their own privacy-related preferences.
A multimodal prediction model which mixes visual content and tags is introduced in~\cite{tonge2019}.
Performance is improved by exploiting predictions from neighboring photos from the user's stream.
These approaches are relevant insofar they focus on improving users' control over shared data by predicting the privacy for individual images.
Our approach is different because feedback is proposed both at user profile and image levels.
Equally important, it is linked to situations to better model real usage of shared data.

Image analysis is central in our study because it extracts actionable information from shared photos.
A choice between deep learning-based image classification and detection is needed. 
Classification~\cite{DBLP:conf/cvpr/HeZRS16,krizhevsky2012imagenet} provides global labels for each image, while detection~\cite{DBLP:conf/iccv/LinGGHD17,redmon2017yolo9000} delineates image regions which contain specific objects.
Detection is better suited in our work since useful information is most often conveyed by localizable objects, which might be missed in classification.
Object detection witnessed the proposal of increasingly accurate methods~\cite{he2017mask,redmon2017yolo9000,redmon2018,ren2015faster}.
However, the most accurate models are often too complex for edge computing.
This is important insofar one objective here is to inform users about the potential effects of sharing before it is done on their smartphones.
More compact models which search for a trade-off between performance and complexity were proposed in~\cite{ghiasi2019,qin2019thundernet,sandler2018}.
Consequently, we compare models which are usable on smartphones and are either generic~\cite{ren2015faster} or specifically designed for edge computing~\cite{sandler2018}.
Note that the use of other detection models could further boost reported performance.

\section{Proposed method}
\label{sec:method}
\subsection{Preliminary experiment}
We hypothesize that linking feedback about the effects of personal data sharing to real-life situations improves its efficiency compared to existing approaches. 
We designed an experiment to test this hypothesis using four real-life situations: bank loan, accommodation search, IT job search and waiter job search.
A set of 20 photos that depict objects with potentially negative effects is selected.
Objects with different negative ratings for each situation as obtained in Subsection~\ref{subsec:crowd} are kept.
They include: casino, knife, cannabis leaf, rifle, etc.
Participants assume that they are about to share a photo and that an AI-driven assistant advises against sharing. 
They see the following messages which implement:
\begin{itemize}[noitemsep]
\vspace{-0.3em}
    \item \textbf{existing feedback}~\cite{orekondy2017,spyromitrosxioufis2016}: "The app predicts that the image should not be shared because it automatically detected \textit{object X} in the image."
    \item \textbf{proposed feedback}: "You are \textit{in situation Y}. The app predicts that the image should not be shared because it automatically detected \textit{object X} in the image. \textit{Object X} is negatively perceived in \textit{situation Y}."
\end{itemize} 
The rest of the interface is identical.
Each participant sees only one type of message to avoid interferences between feedback types.
Images are presented randomly to avoid ordering effects.
Participants are asked to answer the following question: "Would you follow the advice provided by the app and not share the image?".
They can respond "No", "Maybe" or "Yes".
We encode these three responses as 0, 1 and 2 for results analysis.
There were 50 participants for each of classical and proposed feedback.
The average scores for the existing and proposed feedback are 0.86 and 1.13 respectively, with the corresponding standard deviations being 0.87 and 0.91.
The large standard deviation values are normal since the negativeness of depicted objects is variable.
A t-test with independent samples is applied to the sets of averaged photo scores. 
It shows that the difference is significant with $p < 0.001$.
We can conclude to a strongly increased efficiency of the proposed feedback compared to the classical one.
To instantiate feedback related to real-life situations, we introduce a method which learns to rate the effect of shared data both for individual images and entire user profiles.

\subsection{Notations}
The following notations are used below: an user $U^i$ whose profile is rated; the set of $v$ photos of $U^i$ defined as $\mathcal{P}^i=\{P^i_1, P^i_2, ..., P^i_v\}$ and analyzed automatically to rate the profile; a set of $w$ visual objects $\mathcal{D} = \{O^1, O^2, ..., O^w\}$; an object detector $d(O_k^l,P^i_j)$ which detects the $k^{th}$ occurrence of visual object $O^l$ in the $j^{th}$ photo; a situation model $\mathcal{S}=\{r(O^1), r(O^2), ..., r(O^w)\}$ defined by a set of visual objects rated via crowdsourcing $r$; a set of visual profiles for $x$ users $\mathcal{C} = \{U^1, U^2, ..., U^x\}$ with manual profile ratings $m(U^i)$ collected by crowdsourcing; an automatic profile rating $ur(U^i)$ of $U^i$ in situation $\mathcal{S}$. 

\subsection{Crowdsourcing Situation Ratings}
\label{subsec:crowd}
The interpretation of a object might vary between contexts, and so would the effects of sharing its images.
Situations are modeled by crowdsourcing visual objects ratings.
Impactful situations were selected: accommodation search (ACC below), bank credit demand (BANK), job search as IT engineer (IT) and job search as a waitress/waiter (WAIT).
ACC and BANK are applicable to a large part of the population. 
IT and WAIT are relevant for population segments, but the respective job searches require different profiles.
Detectable objects from the OpenImages~\cite{kuznetsova2018open}, ImageNet~\cite{DBLP:journals/ijcv/RussakovskyDSKS15} and COCO~\cite{lin2014microsoft} datasets were rated to boost detector coverage.
A limitation here is that task-relevant objects are missing and $\mathcal{D}$ could be enriched.

A rating interface is created which includes for each situation: the object name, illustrative thumbnails and a 7-points Likert scale with ratings between -3 (strongly negative influence) to +3 (strongly positive influence).
There were 56 participants in total, with 14 rating sets per situation.
The final rating $r$ is obtained by averaging their contributions. 
The resulting detection dataset $\mathcal{D}$ includes 269 objects with $r \neq 0$ for at least one situation.
Inter-rater agreement, which is important for tasks prone to bias such as the one proposed here, is computed using the average deviation index ($AD$)~\cite{burke1999average}.
The obtained $AD$ varies between $0.48$ for $IT$ and $0.65$ for $WAIT$. These values are well below $AD \leq 1.2$, the maximum acceptable value for a 7-points Likert scale defined in~\cite{burke2002estimating}. 
Beyond rater agreement, the qualification of annotators is important. 
We define expertise as having working experience in BANK, IT and WAIT and having a landlord experience for ACC. 
Depending on the situation, the number of annotators with expertise in each situation was: two for BANK, six for IT, four for WAIT and four for ACC.
Their ratings were generally well aligned with those provided by the other annotators.
For instance, the $AD$ agreement between average scores of experts and of other participants is only $0.17$ for $IT$.
This finding indicates that the obtained ratings are representative. 

The mean object ratings are -0.13 for BANK, 0.03 for ACC, 0.09 for IT and 0.27 for WAIT (standard deviations are 0.68, 0.7, 0.58 and 0.6 respectively).
This illustrates the tendency of participants to be stricter when deciding about a bank loan than elsewhere.
The is intuitive because granting a loan has tangible monetary consequences, which are easily internalized by participants.
Inversely, WAIT, a situation with less serious implications, has the highest rating.

\begin{figure*}[t]
\begin{center}
\includegraphics[width=0.85\linewidth,trim={0cm 0.4cm 0cm 0cm}]{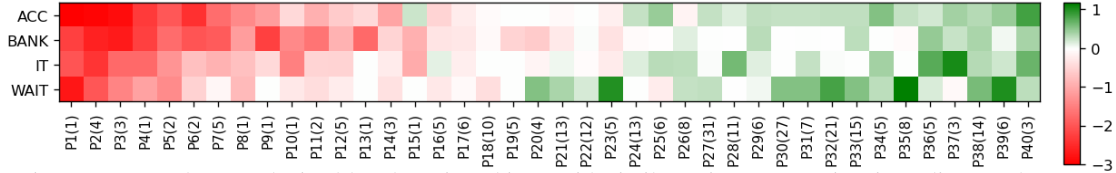}
\caption{Rating patterns (columns) obtained by clustering objects with similar ratings across situations (lines). The pattern name and the corresponding number of objects are provided under each column. Rating colors go from red (strongly negative) to green (strongly positive) with stronger intensity indicating a higher  absolute value of rating. The object details of the patterns are shown in Table 4 of the supplementary material.}
\label{fig:vispatterns}
\end{center}
\vspace{-7mm}
\end{figure*}
A clustering of visual objects based on their rating similarities across situations is shown in Fig.~\ref{fig:vispatterns}.It includes 40 patterns (cluster centroids) discovered by using the lowest silhouette criterion in K-means.
These patterns confirm that object interpretation varies across situations.
Averaged negative ratings are stronger since they reach -3, while averaged positive ratings have a maximum value of 1.15.
Consistent with mean ratings, BANK and ACC have a larger number of negatively rated objects compared to IT and WAIT.
While the negative range is stronger, a majority of objects have low positive scores.
Patterns P27, P30 and P32 include 31, 27 and 21 objects, respectively. 
Some cluster ratings vary strongly between situations. 
P23, P35 and P39 are positive for WAIT but neutral or even negative elsewhere.
These patterns include objects related to junk food and alcohol. 
P1, P2 and P3, which include objects related to weapons, have only negative ratings. 
The relatively low number of strong ratings indicates that the dataset should be enriched with highly rated objects in order to allow a finer-grained computation of visual profile ratings.
More details about objects crowdsourcing are given in the supp. material.

\subsection{Focal rating}
Figure~\ref{fig:vispatterns} indicates that a majority of objects have low ratings and they might overwhelm the less numerous but more significant highly-rated objects. 
Boosting the influence of latter is important and this is done using Eq.~\ref{eq:FE}:
\begin{equation}
    fr(O^l) = \frac{1}{(1-\frac{1}{K}*|r(O^l)|)^{\gamma}}*r(O^l)
    \label{eq:FE}
    \end{equation}
where: $K$ and $\gamma$ control the strength of the focal rating.
This function is inspired by attention mechanisms~\cite{wang2017deep} which were used to improve the performance of deep learning applications, such as object detection~\cite{DBLP:conf/iccv/LinGGHD17}.
Note that $K > |r(O^l)|$ is needed to preserve the sign of $fr$ after scaling with $\gamma$.
$fr$ will have little influence on objects that have low initial $r(O^l)$.
The higher the absolute value of $r(O^l)$ is, the more its effect will be boosted by Eq.~\ref{eq:FE}.
The effect of focal rating is illustrated in the supp. material.

\begin{figure*}[t]
\begin{center}
\includegraphics[width=0.95\linewidth,trim={0cm 0.3cm 0cm 0cm}]{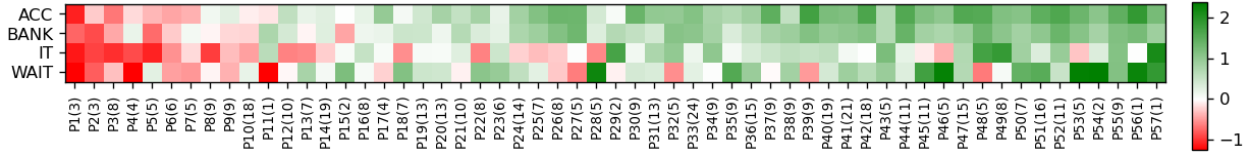}
\caption{Rating patterns (columns) obtained by clustering visual user profiles with similar ratings across situations (lines). The pattern name and the corresponding number of objects are provided under each column. Rating colors go from red (strongly negative) to green (strongly positive) with stronger intensity indicating a higher  absolute value of rating.}
\label{fig:uspatterns}
\end{center}
\vspace{-8mm}
\end{figure*}

\subsection{Crowdsourcing Visual User Profile Ratings}
\label{subsec:profiles}
We collect manual ratings $m(U^i)$ for users $U^i$ in situation $\mathcal{S}$ via crowdsourcing.
Like object rating, visual profiles are evaluated using a 7-point Likert scale that goes from -3 (strongly unappealing) to +3 (strongly appealing).
Ratings are collected from 9 participants for 500 users from the YFCC dataset~\cite{thomee2016yfcc100m} with 100 images per profile.
YFCC was sampled because it includes images that were shared publicly under Creative Commons licenses which allow reuse.
The images of each profile are shown on a single page, along with the possible situation rating.
Participants are asked to look at all the photos and provide a global rating for each user in each situation.
As illustrated in the supplementary material, the evaluated situations are presented in the interface.
Profiles were presented randomly to participants to avoid ordering bias.
Similar to Subsection~\ref{subsec:crowd}, inter-rater agreement is analyzed using the $AD$ index~\cite{burke1999average}.
$AD$ values are $0.86$ for ACC, $0.77$ for BANK, $0.74$ for IT and $0.83$ for WAIT.
These values are within the acceptability bounds defined in~\cite{burke2002estimating} ($AD \leq 1.2$).
There were at least two expert annotators for each situation whose ratings were well aligned with those of other participants. 

A clustering based on the similarity of profile ratings across situations which minimizes silhouette is presented in Fig.~\ref{fig:uspatterns}. 
There are 57 patterns discovered for the 500 profiles in the dataset. 
The ratings of the same users vary significantly from a situation to another. 
Our hypothesis that profile perception is context-dependent is confirmed.
Unlike the object rating patterns from Fig.~\ref{fig:vispatterns}, positive ratings are in the majority here and positive ratings are stronger. 
The most populous patterns are positive, with 24 users in P33, 21 in P41 and 19 in P40.
Most negative patterns include a limited number of profiles.
Patterns such as P26, P27, P28 are rated positively in some situations and negatively in others. 
This finding supports the creation of per-situation rating predictors. 
Due to space limitations, details about profile rating crowdsourcing are given in the supp. material.

\subsection{Baseline Rating of Visual User Profile}
\label{subsec:baseline}
We attempt to obtain a reliable estimation of profile ratings in the modeled situations by using: 
(1) object ratings from Subsection~\ref{subsec:crowd},
(2) object detections proposed by models learned with dataset $\mathcal{D}$ and
(3) the set $\mathcal{C}$ of visual profiles rated in Subsection~\ref{subsec:profiles}.
Given a photo $P^i_j$, detectors search for objects from $\mathcal{D}$ and scores are predicted for all these objects by default.
In practice, a threshold is needed to decide if an object is actually present.
We define:
\begin{equation}
\hat{d_{\eta}}(O_k^l,P^i_j) = 
\begin{cases}
d(O_k^l,P^i_j), & \text{if } d(O_k^l,P^i_j) \geq \eta(O^l)\\
0, & \text{otherwise}
\end{cases}
\label{eq:det}
\end{equation}
where:
$d(O_k^l,P^i_j)$ is the detector for object $O^l$ which evaluates the confidence of the $k^{th}$ raw prediction of $O^l$ in photo $P^i_j$; $\eta(O^l)$ is a detection threshold for $O^l$.

The threshold value $\eta(O^l)$ in Eq.~\ref{eq:det} can be set to a single value for all objects.
This value can be optimized by maximizing the correlation between automatic and manual profile ratings over a validation set $\mathcal{V}$.
However, it does not account for performance variations of object detectors. 
Instead, we implement a filtering-based attribute selection mechanism~\cite{phuong2005choosing} which: (1) determines optimal thresholds for individual detectors and (2) selects only the individual detectors which are most relevant in context.  
We compute:
\vspace{-1mm}
\begin{equation}
(\tau^{m}(O^l),\eta^{m}(O^l)) = \max_{\forall \eta \in [0.01,1]} PR(\mathcal{E}_{\mathcal{V}}^{\eta}(O^l),\mathcal{M}_{\mathcal{V}}^{\eta}(O^l))
\label{eq:th}
\end{equation}
where: 
$\tau^{m}(O^l)$ is the optimal correlation between manual and automatic rankings of profiles obtained using a single $O^l$;
$\eta^{m}(O^l)$ is the detection threshold associated to $\tau^{m}(O^l)$ which ranges from 0.01 to 1 and is tested with a step of 0.01; 
$PR$ is the Pearson correlation coefficient; 
$\mathcal{E}_{\mathcal{V}}^{\eta}(O^l)$
and $\mathcal{M}_{\mathcal{V}}^{\eta}(O^l)$ are the sets of automatic and manual rankings of profile ratings from $\mathcal{V}$ obtained for each $\eta$ tested.

Eq.~\ref{eq:th} optimizes the correlation between automatic and manual user ratings when a single detector is activated.
If several values of $\eta^{m}(O^l)$ maximize $\tau^{m}(O^l)$, the smallest of them is used to keep more occurrences of $O^l$.
The outputs of Eq.~\ref{eq:th} need to be aggregated in order to decide which subset of $\mathcal{D}$ is best in $\mathcal{S}$. 
The optimal $\tau^{m}(\mathcal{D})$ is selected with:
\vspace{-1mm}
\begin{equation}
\tau^{m}(\mathcal{D}) = \argmax_{\forall \tau \in [-1,1]} PR(\mathcal{E}_{\mathcal{V}}^\tau(\mathcal{D}^\tau),\mathcal{M}_{\mathcal{V}}^\tau(\mathcal{D}^\tau))
\label{eq:taumax}
\end{equation}
where: $\tau$ is a correlation value between -1 and 1, with a step of 0.01;
$\mathcal{D}^\tau$ is a subset of $\mathcal{D}$ which activates only a subset of detectors;
and the $PR$, $\mathcal{E}^\tau$ and $\mathcal{M}^\tau$ are the same as in Eq.~\ref{eq:th}.

Eq.~\ref{eq:taumax} provides the correlation threshold which selects the best subset of object detectors $\mathcal{D}^{m}$ for visual profiles in $\mathcal{V}$. 
Individual detectors from $\mathcal{D}$ are activated with:
\vspace{-1mm}
\begin{equation}
\hat{d}^{m}(O_k^l,P^i_j) = 
\begin{cases}
\hat{d_\eta}(O_k^l,P^i_j), & \text{if } \tau^{m}(O^l) \geq \tau^{m}(\mathcal{D})\\
0, & \text{otherwise}
\end{cases}
\label{eq:detmax}
\end{equation}

High values of $\tau^{m}(\mathcal{D})$ create a sparse $\mathcal{D}^{m}$.
Only highly relevant detectors which provide strong correlations $\tau^{m}(O^l)$ for situation $\mathcal{S}$ are activated. 
The disadvantage of using a high $\tau^{m}(\mathcal{D})$ is that the coverage $\mathcal{D}^{m}$ is low, and only a subset of profiles can be reliably characterized.
Inversely, low values of $\tau^{m}(\mathcal{D})$ lead to a dense $\mathcal{D}^{m}$
and ensure good profile coverage at the expense of the relevance of individual detectors.
Eq.~\ref{eq:detmax} finds a balance between the relevance of individual detectors and profile coverage.

The baseline rating of user profiles $ur(U^i)$ in $\mathcal{S}$ combines object ratings $r$ for detections from $\mathcal{D}^{m}$ obtained with Eq.~\ref{eq:detmax}.
It is defined as:
\begin{equation}
ur(U^i) =\frac{\sum_{j=1}^{v} \sum_{l=1}^{w} r(O^l) \sum_{k=0}^{n_l} \hat{d}^{m}(O_k^l,P^i_j)}{\vert\hat{\mathcal{P}}^i\vert}
\label{eq:indiv}
\end{equation}
where:
$n_l$ is the total number of detections of $O^l$ filtered using Eq.~\ref{eq:detmax},
$\vert\hat{\mathcal{P}}^i\vert$ - the cardinality of the subset of $U^i$ photos with at least one visual detector activated.

The denominator $\vert\hat{\mathcal{P}}^i\vert$ produces an averaged profile rating and facilitates the comparison of visual profiles that include a variable number of images.
Assuming that the numerator gives the same result for two users, the rating will be lower for the one which has more images with detections activated.
In this way, profiles that include fewer but more salient detected objects are prioritized.

\subsection{Learning to Rate Visual User Profiles}
\label{subsec:lervup}

The baseline method from Subsection~\ref{subsec:baseline} performs a simple aggregation of available components but does not exploit them fully.
We hypothesize that a supervised learning approach is better suited for profile rating.
$LERVUP$ builds on the baseline and includes an image-level descriptor, a module which makes descriptors more compact at user level, and a model which performs the final prediction.

\vspace{-0.7em}

\subsubsection{Image-level descriptor}
Individual photos are a core factor in the manual rating of user profiles. 
It is thus interesting to aggregate object detections at the image level.
Such a descriptor is equally interesting insofar it provides understandable feedback about individual photo contributions to the profile rating.
The descriptor includes three attributes defined as follows:
\begin{equation}\label{eq:atts}
\begin{split}
\resizebox{1\hsize}{!}{$f_p(P_i^j) =\frac{\sum_{l=1}^{w}r(O^l)\llbracket{r(O^l) \geq 0}\rrbracket\sum_{k=0}^{n_l}\llbracket{\hat{d_\eta}(O_k^l,P_i^j) > 0}\rrbracket}{\sum_{l=1}^w\sum_{k=0}^{n_l}\llbracket{\hat{d}_{\eta}(O_k^l,P_i^j) > 0}\rrbracket}$}\\
\resizebox{1\hsize}{!}{$f_n(P_i^j) =\frac{\sum_{l=1}^{w}r(O^l)\llbracket{r(O^l) < 0}\rrbracket\sum_{k=0}^{n_l}\llbracket{\hat{d}_{\eta}(O_k^l, P_i^j)>0}\rrbracket}{\sum_{l=1}^w\sum_{k=0}^{n_l}\llbracket{\hat{d_{\eta}}(O_k^l, P_i^j)>0}\rrbracket}$}\\
\resizebox{0.75\hsize}{!}{$f_c(P_i^j) = \frac{\sum_{l=1}^w\sum_{k=0}^{n_l}\hat{d_\eta}(O_k^l,P_i^j)}{\sum_{l=1}^w\sum_{k=0}^{n_l}\llbracket{\hat{d}_{\eta}(O_k^l,P_i^j)>0}\rrbracket}$}\\
\end{split}
\end{equation}

where:
$\hat{d}_{\eta}(O_k^l,P_i^j)$ is the filtered confidence of the $k^{th}$ detection with the optimal $\eta$ estimated in Eq. \ref{eq:th} for each $O^l$; $n_l$ is total number of $O^l$ detections; $r(O^l)$ is the $O^l$ rating; $f_p$,$f_n$ and $f_c$ are the positiveness, negativeness and confidence attributes of the image; 
$\llbracket \rrbracket$ is the Iverson bracket, valued 1 if the inner condition is true and zero otherwise. 

$f_p$ and $f_n$ are designed to favor images that include objects with strong impact ratings. 
The higher the absolute values of $r(O^l)$ are on average, the more salient $f_p$ and $f_n$ will be.
$f_c$ gives an average of the valid detection scores from the image and favors images that include high confidence object detections over the others. 
Note that Eq. \ref{eq:atts} is applicable if at least a valid detection exists in the image. 
Otherwise, the image is not considered in $LERVUP$.

\vspace{-0.6em}

\subsubsection{User-level descriptor}
Image-level descriptors is aggregated at user level to mimic the way humans rate visual user profiles.
This is challenging because visual objects with different ratings appear in isolation or jointly in one or several profile images. 
The user-level descriptor aggregation method is described in Algorithm~\ref{user_descriptor}. 
First, we construct $\theta_i$, a set of image descriptors for each user $U^i$. 
The sets $\theta_i$ are aggregated into $F$, which is exploited to train a clustering model $\mathcal{M}$. 
$\mathcal{M}$ is subsequently used to infer \textit{clusters} which group together patterns in the underlying structure of each user profile.
K-means, with $K=4$ clusters, is chosen for its effectiveness and simplicity. 
Other $K$ values were tried but did not improve the obtained results.
The mean and variance of image descriptors from the clusters are concatenated in a final feature vector $f_i$.
It constitutes a better representation of the user profile compared to raw use of object ratings and object detections. 
$f_i$ captures in a compact form patterns from an initial high-dimensional space defined by an array of object detectors and thus avoids the curse of dimensionality~\cite{hughes1968mean}.
The proposed descriptor is an alternative to classical dimensionality reduction techniques~\cite{scholkopf1997kernel, wold1987principal}. 
A comparison to the two forms of compression of raw representations is proposed.

\begin{algorithm}
    \footnotesize
    \DontPrintSemicolon
    \SetKwData{Cluster}{cluster} \SetKwData{Clusters}{clusters}
    \SetKwData{MV}{mv}
    \SetKwFunction{ImageDescriptor}{ImageDescriptor}
    \SetKwFunction{TrainClusteringModel}{TrainClusteringModel}
    \SetKwFunction{MeanVariance}{MeanVariance}
    \SetKwFunction{Concatenate}{Concatenate}
    
    $F \longleftarrow \emptyset$\;
    \For{$U^i$  \textbf{in}  $\mathcal{C}$}{
        $\theta_i \longleftarrow \emptyset$\;
        \For{$P_j^i$ \textbf{in} $\mathcal{P}^i$}{
            $f_p^{ij}, f_n^{ij}, f_c^{ij}$ $\leftarrow$ \ImageDescriptor{$P_j^i$}\;
            $\theta_i \leftarrow \theta_i \cup \{ f_p^{ij}, f_n^{ij}, f_c^{ij} \}$\;
        }
         $F \leftarrow F \cup \theta_i$\;
    }
    $\mathcal{M}$ $\leftarrow$ \TrainClusteringModel{F}\;
    \For{$U^i$  \textbf{in}  $\mathcal{C}$}{
        $f^i \longleftarrow \emptyset$\;
        $\Clusters \leftarrow \mathcal{M}(\theta_i)$\;
        \For{$\Cluster$ \textbf{in} $\Clusters$}{
        $\MV \leftarrow \MeanVariance(\Cluster)$\;
        $f^i \leftarrow \Concatenate(f^i,$\MV$)$\;
        }
    }
   \vspace{-0.5em}
   \caption{\footnotesize User Profile Rating Descriptor}\label{user_descriptor}
\end{algorithm}

\vspace{-0.9em}

\subsubsection{\textbf{\textit{LERVUP}} training}
Visual profile rating is modeled as a regression problem that exploits the user-level descriptor described in  Algorithm~\ref{user_descriptor}.
$LERVUP$ training is deployed as a pipeline process. 
First, individual object detections are validated within each image. 
Second, the image-level descriptor is constructed per image. 
Third, clustering is applied to group together similar image descriptors and discover relevant patterns for the entire training set.
Fourth, the discovered patterns are concatenated to build the user descriptor. 
Finally, a random forest regression model is used to learn the rating of visual user profiles. 
Random forest was chosen because it is robust to data that contain non-linear relationships between features and target variables \cite{kayri2017performance,svetnik2003random,youssef2016landslide}.
Note that the training is scalable since its optimization takes less than an hour per situation on a standard Intel Core $i7$ processor.

\section{Evaluation}
\label{sec:experiments}
The main objective of this first evaluation is to assess the feasibility of the task. 
Note that the user profiles dataset is not large enough to split it into train, validation and test subsets which have sufficient size.
We thus split the dataset in training and validation sets $\mathcal{L}$, and $\mathcal{V}$ which include 400 and 100 profiles, respectively.
The optimal configuration of each method on $\mathcal{V}$ is obtained using grid search and reported below. 
Details about optimized parameters and their ranges are given in the supp. material.
Results are also provided for the ablation of 50\% of user profiles or 50\% visual objects.

\subsection{Object Detection Dataset and Models}
\label{subsec:dataset}
The coverage ensured by the detection dataset is important to enable processing of different types of visual content.
As we mentioned, we merge three existing datasets: OpenImages~\cite{kuznetsova2018open}, ImageNet~\cite{DBLP:journals/ijcv/RussakovskyDSKS15} and COCO~\cite{lin2014microsoft}.
Whenever an object is present in more than one dataset, a balanced sampling is performed.
The resulting dataset includes 269 objects and 137976 images.
We limit imbalance by retaining at most 1000 images per object.
The average and standard deviation of the distribution are 513 and 305, respectively.

Detectors are trained with mobile and generic models.
The mobile model (MOBI) is a MobileNetV2~\cite{sandler2018} with depthwise convolutions, which offer a good precision/speed tradeoff.
The detection head is a Single Shot MultiBox Detector~\cite{DBLP:conf/eccv/LiuAESRFB16}, a fast single-stage method that is adapted for edge computation.
The generic model (RCNN) uses Inception-ResNet-v2~\cite{szegedy2017inception} with atrous convolutions and a Faster RCNN module~\cite{ren2015faster} for detection. 
While not designed specifically for mobile devices, tests showed that it is usable on recent Android smartphones.
Details about detector training are provided in the supp. material.

\subsection{Methods}
We test the following variants of the proposed methods:

\vspace{-0.5em}
\begin{itemize}[itemsep=1pt]
    \item $BASE$ and $BASE_{\eta}$ - ranking based on Eq.~\ref{eq:indiv} with a unique detection threshold and with $\eta^m(O^l)$ optimized per object via Eq.~\ref{eq:th} and object selection from Eq.~\ref{eq:taumax}.
    \vspace{-0.3em}
    \item $BASE_{\eta}^{fr}$ - version of $BASE_{\eta}$ with $fr(O^l)$ (Eq.~\ref{eq:FE}).
    \vspace{-0.3em}
    \item{$REG_{raw}$} and {$REG_{pca}$} - supervised methods using random forest but with the raw features used in $BASE$ and  16-dimensional PCA-compressed feature which offer the best compression performance.
    \vspace{-0.3em}
    \item $LERVUP$ and $LERVUP^{fr}$ - supervised learning method described in Subsection~\ref{subsec:lervup} with $r(O^l)$ and $fr(O^l)$ (Eq.~\ref{eq:FE}) used for object ratings, respectively.
    \vspace{-0.3em}
    \item $LERVUP^{fr}_{U=50\%}$ and $LERVUP^{fr}_{O=50\%}$ - variants of $LERVUP^{fr}$ learned with half of the profiles and half of the objects from $\mathcal{D}$, respectively.
\end{itemize}
\vspace{-0.5em}

\begin{table}
\resizebox{0.48\textwidth}{!}{ 
\centering
\begin{tabular}{|c|c|c|c|c||c|c|c|c|}
\hline
& \multicolumn{4}{c||}{\textbf{RCNN}} & \multicolumn{4}{c|}{\textbf{MOBI}} \\
\hline
& ACC & BANK & IT & WAIT & ACC & BANK & IT & WAIT \\
\hline
$BASE$ & 0.40 & 0.28 & 0.36 & 0.65 & 0.38 & 0.27 & 0.41 & 0.58 \\
\hline
$BASE_{\eta}$ & 0.45 & 0.28 & 0.36 & 0.65 & 0.42 & 0.26 & 0.41 & 0.58 \\
\hline
$BASE_{\eta}^{fr}$ & 0.45 & 0.33 & 0.36 & 0.65 & 0.42 & 0.30 & 0.41 & 0.58 \\
\hline
$REG_{raw}$ & 0.31 & 0.23 & 0.35 & 0.59 & 0.36 & 0.19 & 0.43 & 0.47\\
\hline
$REG_{pca}$ & 0.45 & 0.30 & 0.43 & 0.60 & 0.32 & 0.09 & 0.24 & 0.63\\
\hline
$LERVUP$ & 0.48 & 0.48 & 0.46 & 0.66 & 0.44 & 0.27 & 0.47 & \textbf{0.68} \\
\hline
$LERVUP^{fr}$ & \textbf{0.55} & \textbf{0.50} & \textbf{0.50} & \textbf{0.68} & \textbf{0.49} & \textbf{0.42} & \textbf{0.51} & \textbf{0.68} \\
\hline
$LERVUP^{fr}_{U=50\%}$ & 0.47 & 0.49 & 0.47 & 0.66 & 0.35 & 0.28 & 0.48 & 0.67 \\
\hline
$LERVUP^{fr}_{O=50\%}$ & 0.43 & 0.42 & 0.47 & 0.66 & 0.47 & 0.35 & 0.48 & 0.64 \\
\hline
\end{tabular}
}
\caption{Pearson correlation between automatic and manual rankings of the ratings of visual user profiles. 
Best results in bold.}
\label{tab:results}
\vspace{-3mm}
\end{table}

\vspace{-0.5em}

\subsection{Results}
The performance of the different methods tested is presented in Table~\ref{tab:results}.
Correlations are analyzed using Cohen's interpretation of the Pearson correlation coefficient~\cite{cohen2013statistical}. Correlation is considered weak for values between 0.1 and 0.3, moderate between 0.3 and 0.5 and strong above 0.5.
All evaluated methods provide a positive correlation between manual and automatic rankings of the profile ratings, with a wide majority of reported correlations in the moderate or strong ranges.
This is a first positive result since the evaluated task is a complex one.
Performance variations observed along different axes are discussed below.

The comparison of the two object detectors is globally favorable to RCNN. 
This result is intuitive insofar RCNN is built with a higher capacity deep network architecture. 
WAIT is the easiest situation, with up to 0.68 correlation obtained for both detectors.
This good behavior for WAIT is explained by the fact that MOBI is known to provide good detections for large objects~\cite{sandler2018}.
However, MOBI also has the worst results by a large margin compared to RCNN.
The maximum correlation value obtained with MOBI for BANK is 0.42 while the corresponding value for RCNN is 0.5. 
These results point out that further performance improvement should be achievable with better object detectors.

The best global results are obtained with $LERVUP^{fr}$.
Six out of eight of the correlations provided by this method are in the strong range defined by~\cite{cohen2013statistical}, with the other in the moderate range.
$LERVUP^{fr}$ clearly outperforms the baselines. 
This finding validates the utility of the learning-based approach, which models automatic profile ranking as a regression problem. 
$LERVUP^{fr}$ is also better than $LERVUP$, with up 15 points gained over it (BANK with MOBI detector). 
The boosting of highly-rated objects via the attention mechanism from Eq.~\ref{eq:FE} is thus validated. 

The four modeled situations have variable performance. 
WAIT is the easiest situation (correlation up to 0.68) because the detection dataset contains a large number of food and beverage-related objects, which are often easy to detect. 
WAIT approximates an upper-bound performance one can expect with the available detection dataset.
BANK is the most challenging situation tested, particularly for MOBI, which offers a maximum correlation of 0.42. 
More object detectors are required to improve results for this situation.

Among the baselines, $BASE_\eta^{fr}$ is better than $BASE_\eta$ and $BASE$.
The introduction of focal rating has a smaller effect on the baseline compared to $LERVUP$.
Object selection (Eq.~\ref{eq:taumax}) and the use of individual detection thresholds (Eq.~\ref{eq:th}) have a larger impact than focal rating. 

Results for $REG_{raw}$ and $REG_{pca}$ are particularly interesting since these methods exploit a random forest training like $LERVUP$.
The difference is that raw user representations, complete or compressed, are fed into $REG$, while $LERVUP$ exploits the proposed compact descriptor.
The difference of performance in favor of $LERVUP$ validates the relevance of the proposed descriptor.

To test the influence of the amount of training data, we ablate half of the user profiles in $LERVUP^{fr}_{U=50\%}$ and half of the visual objects in $LERVUP^{fr}_{O=50\%}$, respectively. 
The comparison of results obtained with the full training set ($LERVUP^{fr}$) to the ablated versions confirms that additional data is clearly beneficial.
Interestingly, removing objects has a stronger effect on RCNN results, while removing users has more impact on MOBI results.
An explanation for this finding is that RCNN detection is more precise than MOBI and thus benefits more from the availability of a larger dataset.
The ablation results have practical significance since the more data are clearly useful.
Note that the effort needed to add the profiles and objects is significant and datasets enrichment is left for future work.

\vspace{-0.7em}

\subsubsection{\textbf{\textit{LERVUP}} implementation}
\begin{figure}
\begin{center}
\includegraphics[width=0.48\textwidth,trim={0cm 0cm 0cm 0cm}]{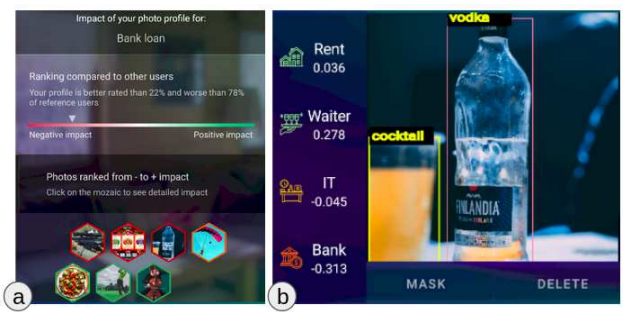}
\caption{Illustration of $LERVUP$-based feedback integrated in a mobile app at profile level (a) and photo level (b).}
\label{fig:feedback}
\end{center}
\vspace{-8mm}
\end{figure}

We describe an implementation of the proposed approach in an Android app which raises users' awareness~\footnote{\url{https://ydsyo.app/}}.
It provides feedback about the effect of image sharing in the modeled situations.
The app is designed to enforce user privacy and the entire processing is done on the user's device.
It is possible to create a profile made of local images, of images already shared on social media, or a combination of them. 
We note that user assistance is most effective for local images than for those which were already shared. 
A traffic light coding of the feedback (from red - negative to green - positive)~\cite{anderson1994minimising} is used in the app to facilitate its understanding. 
Fig.~\ref{fig:feedback} summarizes profile-level feedback (a) and photo-level feedback (b).
Individual profile ratings are difficult to interpret in isolation.
A reference community is created by sampling 5000 visual YFCC profiles which are processed with $LERVUP$.
The target profile rating is provided by comparison to the reference community. 
The same page includes thumbnails for images ranked by their impact in the situation.
The user can select any photo for a detailed view of its effects (Fig.~\ref{fig:feedback} (b)).
Feedback is based on the photo impact rating given by Eq. ~\ref{eq:atts} and is color-coded for each situation.
The detected objects are highlighted using bounding boxes. 
The same screen contains two buttons that instantiate a control mechanism for local photos.
"DELETE" removes the photo from the device.
"MASK" changes its visibility to avoid uploads to social networks.
Such control is desirable also for photos which were already shared but is not permitted by existing APIs.

\vspace{-0.1em}

\section{Discussion and Conclusion}
We presented a new approach that unveils potential real-life effects of photo sharing. 
It is implemented for four situations but is extensible in terms of situations, types of data included, object detection models and profile rating methods.
Below, we discuss the implications of the work, point out its merits, limitations and perspectives of improvement.

The main objective is to help users understand how their shared data could reflect on their real lives. 
While promising, the approach is affected by a combination of human and technical biases.
However, such biases are inherent to any AI-driven computer system and will also appear in real decision-making processes, which are mimicked here.
For instance, crowdsourced object and profile ratings will reflect an average bias of the participants who provided inputs in the experiments.
In a real context, where shared data is analyzed by a single person to reach a decision, the bias will be personal.
Averaging ratings is a good way to reduce bias since any extreme individual opinions will be smoothed. 
In the future, it would be interesting to collect data from a larger pool of participants and cluster them in order to see if there are large rating differences between sub-communities.
It is also important to note that we encode both positive and negative human biases since participants are asked to rate objects and profiles on a symmetric scale. 
This approach is more balanced than that of studies whose objective is to elicit negative biases~\cite{diazferreyra:hal-01677149, DBLP:journals/ijsc/FriedlandC11}.
Ideally, decision-making processes should be bias-free but it is realistic to assume that biases cannot be eliminated.
It is thus important to act toward at least removing the most damaging of them, which are related to sensitive demographics such as ethnicity, religion or gender.
This topic will be part of our future work.

Technical biases are due to imperfections in the detection model, the available data and the $LERVUP$ model.
Detection model imperfection can be reduced via the use of more powerful deep detection architectures~\cite{DBLP:conf/iccv/LinGGHD17,qin2019thundernet}.
However, since the rating is most useful if done on the users' devices, models should remain tractable at the edge.
A second technical bias is due to detector availability.
Three existing datasets are merged to improve detector coverage.
They seem sufficient for WAIT, which is well mapped in the detection dataset, but probably not for the other situations.
We will extend the dataset with priority given to new objects which are highly rated in at least one situation.
A third technical bias is due to data imbalance.
We limited the maximum number of images per object to reduce imbalance while also preserving accuracy.
The imbalance will be further reduced with new annotations for both existing and new detectors.
A fourth bias is related to the focus on images.
The approach is extensible to other relevant data types, such as likes and texts studied in~\cite{Kosinski5802} and~\cite{petkos2015}.
We intend to exploit them in order to obtain more relevant and broader profile ratings.
Finally, $LERVUP$ provides performance gains compared to a series of baselines.
The proposed method constitutes a first attempt to tackle profile rating, and important improvements over it are possible.

$LERVUP$ is implemented to mimic real decision-making processes and make them more transparent.
A key challenge regarding transparency is related to the explainability of the decision-making process.
The experiment presented in Section~\ref{sec:method} indicates that the high-level explanation of situation-based feedback is more efficient than existing feedback.
Future work will focus on: (1) adding sufficient profiles to the dataset in order to obtain train, validation and test splits which are large enough and (2) improving the explainability of learned models, especially that of deep object detectors, which is the most challenging~\cite{wolf2019explainability}.
The rest of the processing is easier to explain from a technical perspective but still requires a fair amount of AI-related knowledge.

The code is available~\footnote{\url{https://github.com/v18nguye/lervup_official}}.
An anonymized and modified version of the dataset is provided~\footnote{\url{https://www.aicrowd.com/challenges/imageclef-2021-aware/}}.

\vspace{5mm}

\textbf{Acknowledgment}
This work was supported by the European Commission under European Horizon 2020 Programme, grant number 951911 - AI4Media.
This work was supported by the Fondation MAIF.
It was made possible by the use of the FactoryIA supercomputer, financially supported by the Ile-de-France Regional Council.

{\small
\bibliographystyle{ieee_fullname}
\bibliography{ms}
}

\end{document}


\title{Supplementary material for "Unveiling Real-Life Effects of Online Photo Sharing"}

\author{Van-Khoa Nguyen, Adrian Popescu, Jérôme Deshayes-Chossart\\
Université Paris-Saclay, CEA, List, F-91120, Palaiseau, France\\
{\tt\small khoa.v18nguyen@gmail.com,\{adrian.popescu,jerome.deshayes-chossart\}@cea.fr}
}

\maketitle
\thispagestyle{empty}

\section{Introduction}


In this supplementary material, we provide details about:
\begin{itemize}
    \item The experiment which compares classical and proposed feedback.
    \item The crowdsourcing process used to collect visual concept ratings for the modeled situations.
    \item The crowdsourcing process used to collect visual user profile ratings for the modeled situations.
    \item The effect of the focal rating component.
    \item The optimization of $LERVUP$ training.
    \item The optimization of detection models training.
\end{itemize}

\begin{figure*}[t]
\begin{center}
\includegraphics[width=0.97\textwidth]{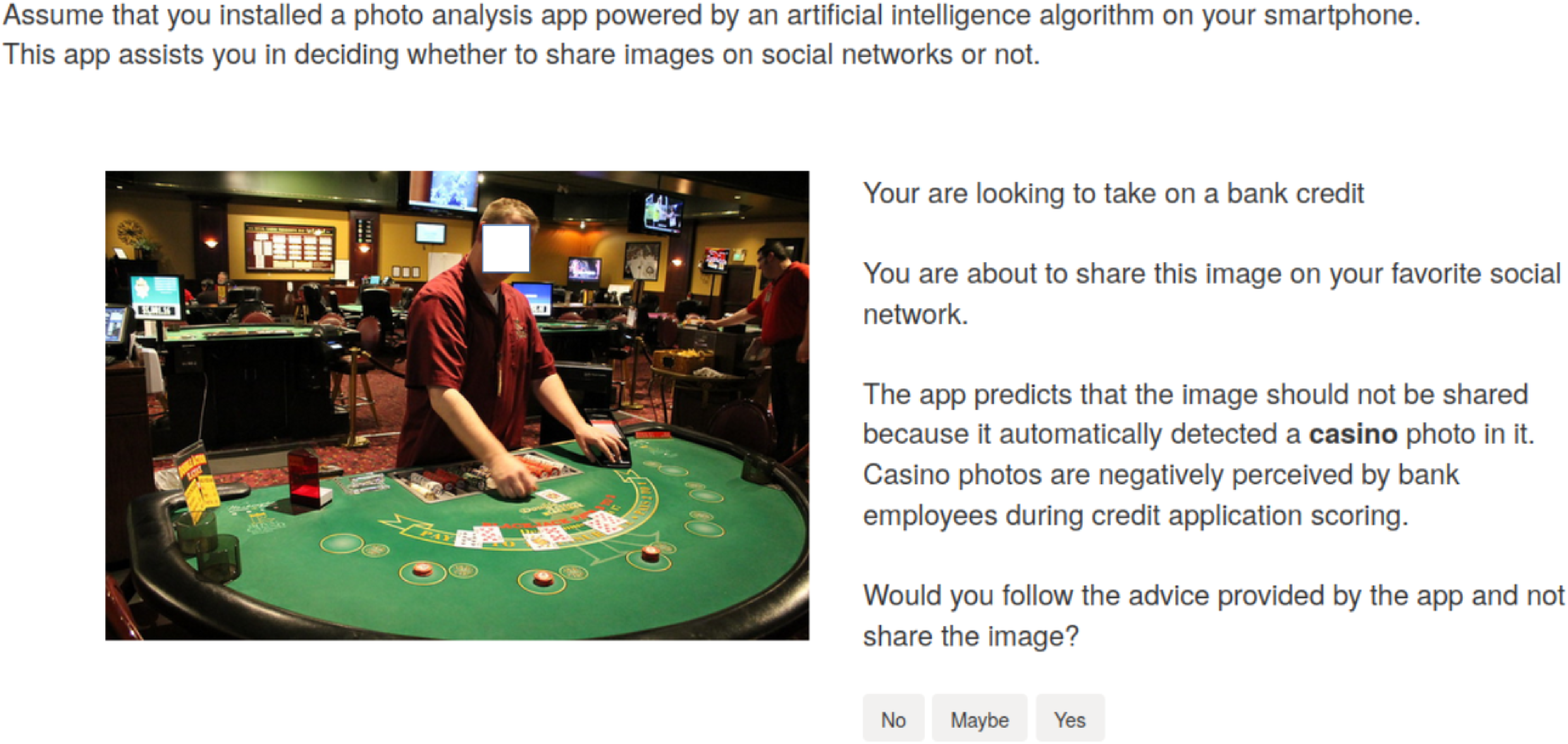}
\caption{Interface used to test the efficiency of classical and proposed feedback. The interface is illustrated for the proposed feedback. It is identical in the two cases with the exception of the message presented as feedback.}
\label{fig:motivation}
\end{center}
\end{figure*}

\section{Feedback comparison experiment}

\subsection{Task and interface}
The description of the task provided to participants at the beginning of the experiment is the following:

\textit{We need your help to optimize the development of an app powered by artificial intelligence whose objective is to assist you when sharing photos on social networks.
You will see one photo per page and should assume that you want to share it on your favorite social network.
The app estimates that the photo should not be shared.
Your task is to decide if you will go on with the sharing or not, depending on the advice provided by the app.
The task includes a total of 20 photos and should not take more than 5 minutes.}

The interface used in the task is illustrated in Figure~\ref{fig:motivation} with an example for the proposed feedback.

\subsection{Participant demographics}
The total number of participants was 100, where 50 of them rated the classical feedback and the others rated the proposed feedback.
27 participants were female and 73 were male.
Their mean age is 28 years old, with the youngest being 18 and the oldest being 47.

\begin{table}
\resizebox{0.48\textwidth}{!}{ 
\centering
\begin{tabular}{|c|c|c|c|}
\hline
\multicolumn{2}{|c|}{\textbf{classical feedback}} & \multicolumn{2}{|c|}{\textbf{proposed feedback}} \\
\hline
concept & avg. score & concept & avg. score \\
\hline
joint (IT) & 1.54 & joint (IT) & 1.76 \\
rifle (WAIT) & 1.4 & bullet (ACC) & 1.68 \\
cannabis\_leaf (ACC) & 1.28 & rifle (WAIT) & 1.58 \\
cigarette\_pack (BANK) & 1.26 & cannabis\_leaf (ACC) & 1.56 \\
bullet (ACC) & 1.2 & knife (BANK) & 1.54 \\
knife (BANK) & 1.12 & casino (BANK) & 1.45 \\
missile (IT) & 1.1 & slot (ACC) & 1.42 \\
pistol (IT) & 1.06 & missile (IT) & 1.32 \\
chicha (WAIT) & 1.02 & chicha (WAIT) & 1.22 \\
axe (ACC) & 0.94 & axe (ACC) & 1.2 \\
demonstration (IT) & 0.84 & vodka (BANK) & 1.2 \\
vodka (BANK) & 0.74 & cigarette\_pack (BANK) & 1.12 \\
tank (ACC) & 0.72 & pistol (IT) & 1.08 \\
slot (ACC) & 0.72 & tank (ACC) & 0.97 \\
casino (BANK) & 0.48 & demonstration (IT) & 0.94 \\
doll (WAIT) & 0.42 & bikini (BANK) & 0.57 \\
cocktail (IT) & 0.4 & cocktail (IT) & 0.54 \\
fighter (WAIT) & 0.34 & baby (WAIT) & 0.52 \\
baby (WAIT) & 0.34 & doll (WAIT) & 0.48 \\
bikini (BANK) & 0.32 & fighter (WAIT) & 0.38 \\
\hline
\end{tabular}
}
\caption{Ranked lists of concepts, with associated situations, presented during the experiment for classical and proposed feedback. The larger the average score is, the more likely it is that the users change their mind and do not share the photo which depicts the concept. Situation codes are as follows: ACC - accommodation search, BANK - bank credit search, IT - IT job search, WAIT - waiter/waitress job search.}
\label{tab:concepts}
\end{table}

\subsection{Discussion of results}
We detail the discussion of results from the beginning of Section 3 of the paper here.
Table~\ref{tab:concepts} lists the averaged scores of the concepts illustrated in the photos for the two types of feedback tested.
To verify whether the two ranked sets of concepts are coherent, we compute the Pearson correlation between them.
The correlation value is 0.82, which indicates that the two sets are strongly linked.
This consistency between the rankings is a positive result insofar the results obtained with the two types of feedback are consistent.  
Intuitively, concepts that have the most negative impact (see Table~\ref{tab:top_pos_neg}) are those for which users will most likely change their minds and not share the corresponding photos.

\begin{table}

\centering
\begin{tabular}{|c|c|c|c|c|}
\hline
Feedback type & ACC & BANK & IT & WAIT \\
\hline 
Classical &  0.97 & 0.78 & 0.98 & 0.70 \\
\hline
Proposed & 1.36 & 1.18 & 1.12 & 0.84 \\
\hline
\end{tabular}

\caption{Average scores per situation with classical and proposed feedback.}
\label{tab:sit_impact}
\end{table}

We also compute aggregated average scores per situation to assess how impactful each situation is and present results in Table~\ref{tab:sit_impact}.
Note that there are only five concepts per situation that are presented and, although they have variable impact scores, the sample remains relatively small to obtain full comparability.
Interestingly, the order of situations changes from classical to proposed feedback.
ACC and IT get nearly the same scores with classical feedback but providing feedback about accommodation seems to be more impactful with the proposed method.
BANK is ranked lower than IT for classical feedback, but the scores are reversed for the proposed feedback.
One explanation for this finding is related to the fact that ACC and BANK are situations that apply to most participants.
They are thus more likely to understand the impact of photo sharing in these situations.

\begin{figure*}[t]
\begin{center}
\includegraphics[width=0.97\textwidth]{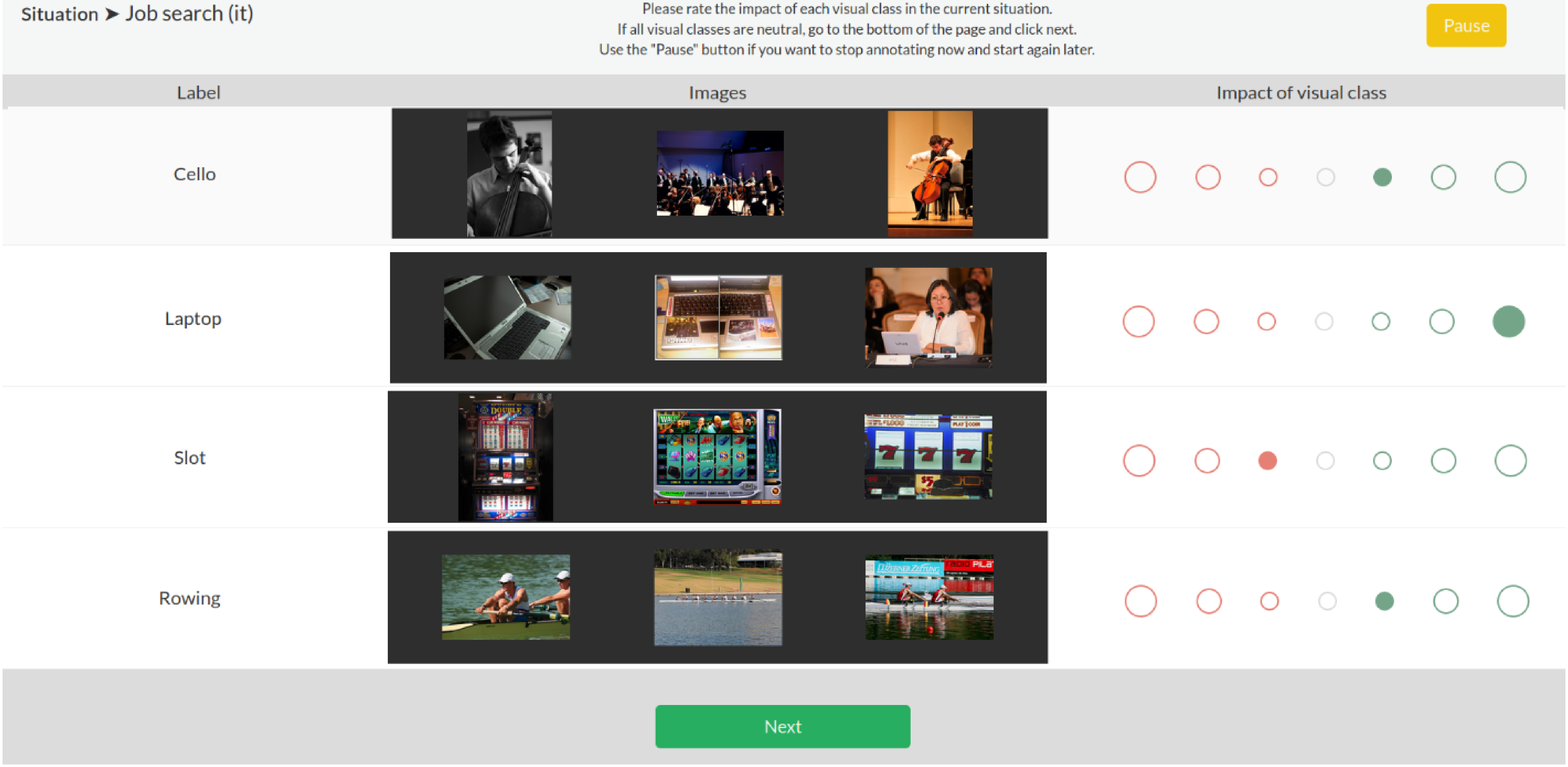}
\caption{Interface used to rate visual classes. The situation name is given in the top-left corner. A short reminder of the task is provided in the middle of the top row. The name, three relevant images and the available ratings are provided for each class.}
\label{fig:classes}
\end{center}
\end{figure*}

\section{Crowdsourcing visual concept ratings}

\subsection{Task and interface}
The description of the task provided to the participants at the beginning of the visual class rating process in the following:

\textit{
We need your help to annotate content for a mobile app which will provide feedback about the effects of online photo sharing to its users.
Your contribution will be used to evaluate the performance of the artificial intelligence tools which will power the mobile application.
Focus is put on photo sharing since images constitute a large part of the content shared online. 
To make the effects easily understandable, we propose a modeling of the impact of data sharing in real-life situations. 
Your will assume that you need to rate visual concepts with respect to their influence when a decision is made about hiring someone as an information technology engineer.
Ratings are provided on a seven points scale and range from "strongly negative influence" to "strongly positive influence". 
}

The interface used for collecting the ratings of visual concepts is presented in Figure~\ref{fig:classes}.

\subsection{Participant demographics}
The task was completed by a total of 52 participants. 20 of them were female and 32 were male. 
Their mean age is 32 year old, with the youngest and the oldest being 23 and 49, respectively. They came from the following countries: France (20), Vietnam (10), Romania (7), Algeria (7),Germany (5), Italy (2), Morocco (1).

\subsection{Examples of obtained results}

\begin{table*}\centering
\begin{center}
\resizebox{0.99\textwidth}{!}{
\begin{tabular}{|c|c|c|}
\hline
\textbf{Situation} & & \textbf{Visual objects with highest positive/negative rating} \\
\hline
\multirow{2}{*}{ACCOM} &  + &  book (1.23); bookshop (1.15); trimaran (1.08); palm tree (0.92); sea lion (0.92); houseplant (0.92); castle (0.92); canoe (0.84); mountain bike (0.83); sunflower (0.76) \\ 
\cline{2-3} 
& - & cocaine (-2.84); bullet (-2.61); revolver (-2.61); pistol (-2.61); rifle (-2.53); machine gun (-2.53); weed (-2.53); joint (-2.30); cannabis leaf (-2.23); cigarette pack (-2.15)\\ 
\hline

\multirow{2}{*}{BANK} &  + &  mountain bike (0.90); cello (0.83); book (0.83); bow tie (0.76); tennis (0.76); harp (0.75); golf (0.69); castle (0.66); strawberry (0.66); salmon (0.66) \\ 
\cline{2-3} 
& - & weed (-2.75); revolver (-2.61); machine gun (-2.53); bullet (-2.46); rifle (-2.38); pistol (-2.38); joint (-2.33); cocaine (-2.25); cannabis leaf (-2.08); dice (-2.08)\\ 
\hline

\multirow{2}{*}{IT} &  + &  book (1.5); web site (1.27); notebook (1.27); laptop (1.18); violin (1.0); tennis (1.0); bicycle (1.0); cello (0.91); piano (0.91); volleyball (0.91) \\ 
\cline{2-3} 
& - & machine gun (-2.33); pistol (-2.33); revolver (-2.25); cocaine (-2.08); bullet (-2.0); rifle (-2.0); joint (-1.75); cannabis leaf (-1.75); stiletto knife (-1.58); weed (-1.58)\\ 
\hline

\multirow{2}{*}{WAIT} &  + &  parfait (1.36); red wine (1.27); trifle (1.18); eggnog (1.18); sidecar (1.18); rugby (1.09); tiramisu (1.09); brew (1.09); tart (1.0); cappuccino (1.0) \\ 
\cline{2-3} 
& - & cocaine (-2.45); revolver (-2.18); pistol (-2.18); bullet (-2.09); machine gun (-1.90); rifle (-1.81); stiletto knife (-1.63); scimitar (-1.36); weed (-1.36); joint (-1.27)\\ 
\hline

\end{tabular}
}
\end{center}
\caption{Top 10 visual positively and negatively rated objects and their ratings for each modeled situation. 
}
\label{tab:top_pos_neg}
\vspace{0mm}
\end{table*}

Table~\ref{tab:top_pos_neg} lists the top 10 positively and negatively rated objects for each situation. 
Results are intuitive, with high and low ratings associated with objects from categories which are positively and negatively connoted.
Negatively rated objects generally belong to categories such as weapons (machine gun, revolver, rifle) or drugs (cocaine, cannabis leaf or weed).
The order of appearance of negatively rated objects varies depending on the situation.
The most salient positive categories are more situation-dependent.
Cultural artifacts, outdoor activities and living entities are salient for ACCOM. 
Sports, cultural artifacts and healthy food are positively connoted for BANK. 
Computing-related artifacts, sports and cultural artifacts are salient for IT. 
Food and drinks, irrespective of them being healthy or not, constitute a large majority of most positively rated objects for WAIT.


\begin{table*}[ht]
\begin{center}
\begin{adjustbox}{width=1\textwidth,center=\textwidth}
\begin{tabular}{ |c|c|c|c| }
\hline
\textbf{Pattern} &\textbf{Related visual concepts} \\
\hline
P40& canoe, ski, book \\
\hline
P39& tea, dessert, tart, orange juice, tiramisu, salad \\
\hline
P38& guitar, piano, cello, violin, harp, flute, snorkel, trumpet, trombone, mango, guacamole, orange, peach, fruit salad \\
\hline
P37& bicycle, mountain bike, notebook \\
\hline
P36& tennis, paddle, golf, volleyball, rowing \\
\hline
P35& goblet, pizza, cappuccino, parfait, hot pot, shortcake, cookie, trifle \\
\hline
P34& dinosaur, sculpture, sea lion, rugby, horse \\
\hline
P33& \multirow{2}{64em}{strawberry, watermelon, cress, hip, honey, cauliflower, gazpacho, apple, cucumber, spinach, artichoke, salmon, lettuce, bok choy, granny smith} \\
& \\
\hline
P32& \multirow{2}{64em}{lobster, cabbage, pretzel, pastry, pasta, pumpkin, spaghetti, milk, cake, rambutan, pancake, dough, omelet, water bottle, fish, pudding, meat loaf, congee, ice cream, espresso, flan} \\
& \\
\hline
P31& coffee cup, pineapple, tomato, grape, waffle, grapefruit, croissant \\
\hline
P30& \multirow{3}{64em}{chestnut, zucchini, bread, mushroom, durian, parsnip, gyro, asparagus, carbonara, butternut squash, broccoli, jalapeno, carrot, quince, pear, guava, marinara, jackfruit, corn, shrimp, fig, persimmon, acorn, chili, fancy dress, muffin, nacho} \\
& \\
& \\
\hline
P29& tie, dogsled, palm tree, suitcase, pomegranate, golf cart \\
\hline
P28& surfboard, boat, backpack, bookshop, calculator, headset, drone, baseball, stick, nursery, web site \\
\hline
P27& \multirow{3}{64em}{cherry, gown, record player, synagogue, totem pole, church, cathedral, sunflower, bakery, banana, lemon, porridge, spoon, oatmeal, buckeye, crab, sushi, bell pepper, sandwich, coconut, western, soup, rapeseed, potato, chard, chowder, celery, bagel, apron, supermarket, hot dog} \\
& \\
& \\
\hline
P26& microphone, castle, drum, wicket, stamp, liner, motorcycle, skateboard \\
\hline
P25& gondola, houseplant, trimaran, crib, unicycle, wheelchair \\
\hline
P24& laptop, dome, flower, jam, egg, trench coat, roller skates, boomerang, segway, bow, oyster, golf ball, skull \\
\hline
P23& cocktail, eggnog, brew, wine, sidecar \\
\hline
P22& bow tie, mosque, sombrero, printer, clipper, tricycle, seat belt, shovel, press, barbecue, burrito, perfume \\
\hline
P21& french fries, sarong, champagne, wine glass, beer, bottle opener, canteen, popcorn, candy, hamburger, doughnut, maillot, jockey \\
\hline
P20& corkscrew, red wine, wine bottle, cheeseburger \\
\hline
P19& ipod, toilet, aircraft carrier, interceptor, limousine \\
\hline
P18& jet ski, cannon, fedora, speedboat, bonnet, parachute, doll, chisel, police boat, motorboat \\
\hline
P17& scissors, chain saw, christmas tree, jeep, fighter, ambulance \\
\hline
P16& diaper, hammer, drill, stretcher, collar \\
\hline
P15& bikini \\
\hline
P14& bob marley, hatchet, ax \\
\hline
P13& card \\
\hline
P12& saw, banner, missile, demonstration, police van \\
\hline
P11& knife, e cig \\
\hline
P10& nude body \\
\hline
P9& dice \\
\hline
P8& sword \\
\hline
P7& tank, chicha, slot, scratch card, casino \\
\hline
P6& cigarette pack, cigarette \\
\hline
P5& scimitar, stiletto knife \\
\hline
P4& cannabis leaf \\
\hline
P3& joint, rifle, weed \\
\hline
P2& pistol, bullet, machine gun, revolver \\
\hline
P1& cocain \\
\hline
\end{tabular}
\end{adjustbox}
\caption{Details of visual concepts contained in the discovered patterns by the K-means algorithm. The patterns are ranged from the most positive pattern (P40) across the four studied situations to the most negative one (P1).}\vspace{1ex}
\label{tab:patterns}
\end{center}
\end{table*}

Table~\ref{tab:patterns} provides the detailed view of the rating patterns obtained for visual concepts which were summarized in Figure 1 of the main paper. 
Patterns are ranked from most positive to most negative average score obtained for the four modeled situations.  
Positive patterns are notably related to sports, music and healthy food. 
Negative patterns are often related to unhealthy habits, drugs and weapons. 
Results from Table~\ref{tab:patterns} are coherent with those presented in Table~\ref{tab:top_pos_neg} but they present a smoothed version of ratings due to the clustering-based pattern creation.

\begin{figure*}[t]
\begin{center}
\includegraphics[width=0.97\textwidth]{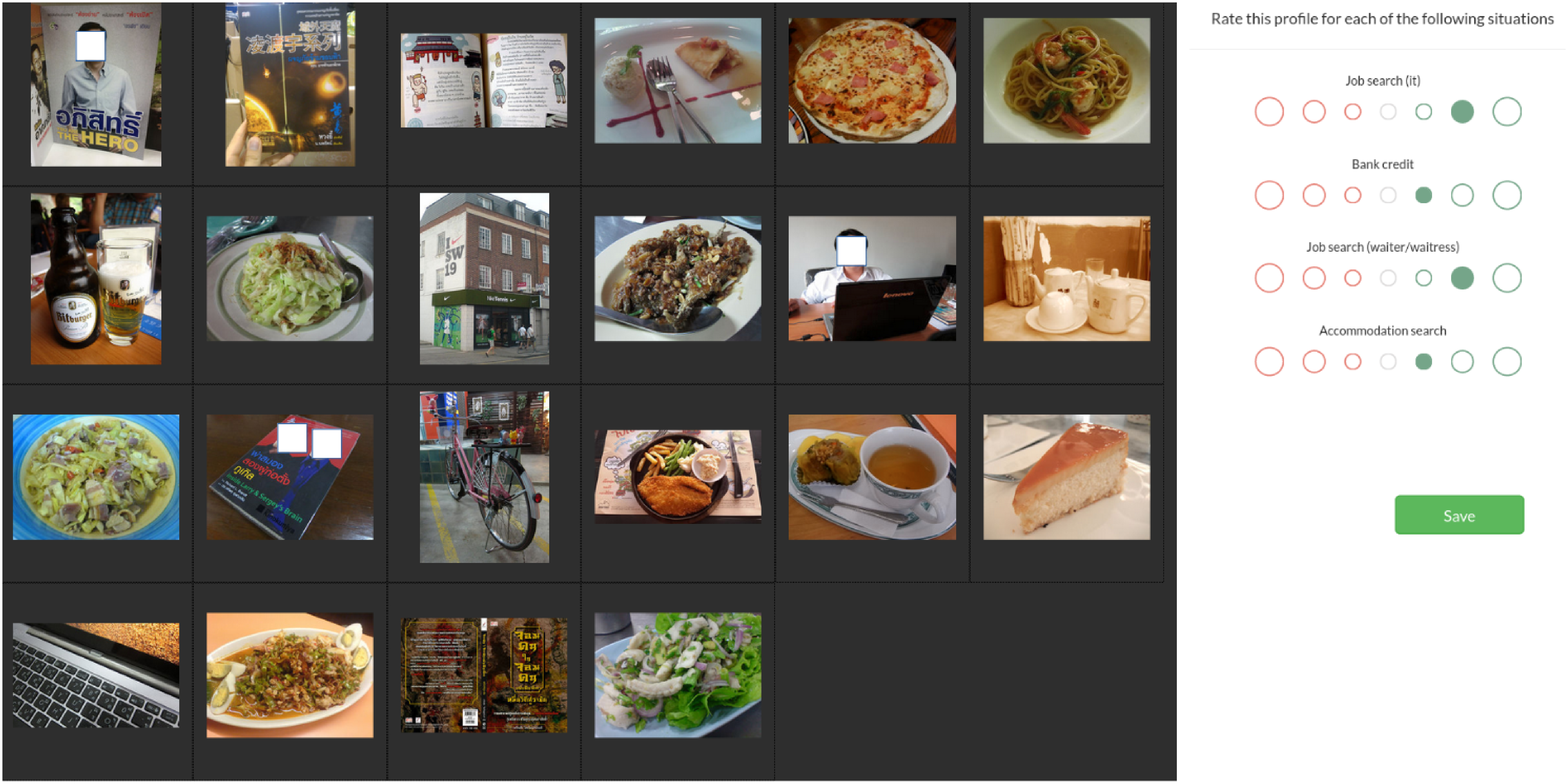}
\caption{Interface used to rate visual user profiles. Only a subset of the 100 photos are presented for the profile. Faces are removed to ensure anonymity.}
\label{fig:profiles}
\end{center}
\end{figure*}

\section{Crowdsourcing visual user profile ratings}

\subsection{Task and interface}

The description of the task provided to participants at the beginning of the rating process was the following:

\textit{We need your help to annotate content for a mobile app which will provide feedback about the effects of online photo sharing to its users.
Your contribution will be used to evaluate the performance of the artificial intelligence tools which will power the mobile application.
Focus is put on photo sharing since images constitute a large part of the content shared online. 
To make the effects easily understandable, we propose a modeling of the impact of data sharing in real-life situations. 
Your input is needed to provide a global rating of a user's visual profile in a situation which has an impact on that user’s real-life.
Each visual profile includes 100 photos published by the same user.
You should browse through the entire set of images until you feel confident about the rating associated to each situation.
You will assume that you are in a decision making position to:
\begin{enumerate}        
       \item Hire the user as an information technology engineer,
       \item Evaluate the user’s bank credit worthiness,
       \item Hire the user as a bartender/waiter,
       \item Rent a flat to the user
\end{enumerate}
Ratings are provided on a seven points scale and range from "strongly repelling profile" to "strongly appealing profile".
}

After reading this description of the task, each participant went through a training session whose objective was to discuss task-related questions with the experimenter. 
Then, participants provided basic demographic information and started the actual task.
The interface used for collecting visual profile ratings is presented in Figure~\ref{fig:profiles} with a subset of photos for one of the users included in the dataset.

\subsection{Participant demographics}
The task was completed by a total of nine participants. 
Four of them were female and five were male.
Their mean age is 34 years old, with the youngest and the oldest being 23 and 47 years old respectively.
They came from the following countries: France (3), Romania (3), Vietnam (2), Colombia (1).

\subsection{Examples of obtained results}
\begin{figure*}
\centering

\includegraphics[width=.999\textwidth,trim={0cm 0cm 0cm 0cm}]{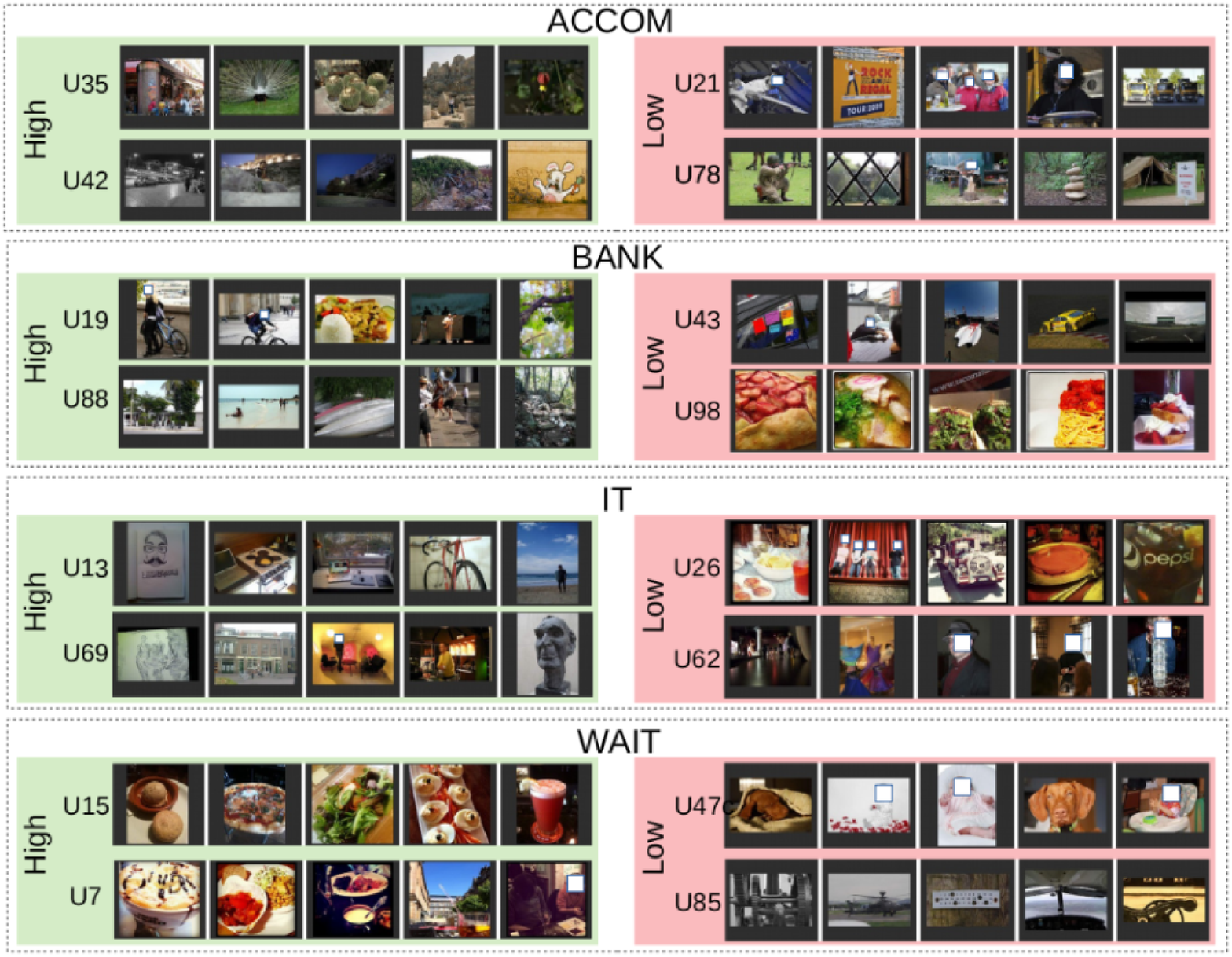}
\caption{Summaries of visual profiles with high and respectively low ratings in each situation.}
\label{fig:samples_profiles}
\vspace{0mm}
\end{figure*}

We illustrate the visual profile ratings with two high and low ranked users for each situation in Figure~\ref{fig:samples_profiles}. 
Five representative images were selected to create visual profile summaries.
The images of the presented profiles are relatively well correlated with the positively and negatively rated objects from Table~\ref{tab:top_pos_neg}.
This correlation points toward a degree of coherence between the ratings provided in the two experiments.
Consequently, the matching of manual and automatic user exposures seems doable.
A detailed look at the profiles from Figure~\ref{fig:samples_profiles} shows that highly rated users for ACCOM shared images of cultural artifacts and of nature. 
Low rated profiles for this situation include images which indicate an inclination for partying (U21) and with a military theme (U78). 
U78 is a very interesting example because the images with a military theme are clearly from historic reenactments but still have negative influence on the profile rating.
Highly rated profiles for BANK depict healthy lifestyles, while low rated ones are linked to motor sports (U43), a risky activity, and unhealthy food (U98).
For IT, highly rated profiles notably include computing-related objects (U13) and cultural artifacts (U69).
U26 and U62 have low ratings because they shared images of drinks, albeit not necessarily of alcohol.
Highly rated profiles for WAIT depict food and drinks, thus showing an interest for objects which are important in this context.
Low rating of U47 for WAIT is linked to sharing images of babies. 
Such photos indicate that this user might be unavailable to work with a variable schedule and at night, a flexibility which is necessary for waiters.
This result is at odds with the one about family structure not influencing job seeking ~\cite{acquisti2020}, probably because these authors did not use photos to characterize user profiles.
U85 notably includes images with a military theme which are again rated negatively. 
Some of the signals illustrated in Figure~\ref{fig:samples_profiles} seem weak out of context and relatively not harmful.
However, they take a negative connotation when interpreted in real-life situation, especially if the interpretation is performed by a machine.
This is notably the case for partying images for ACCOM, motor sports for BANK and baby images for WAIT.

\begin{figure}[t]
\begin{center}
\includegraphics[width=0.99\linewidth]{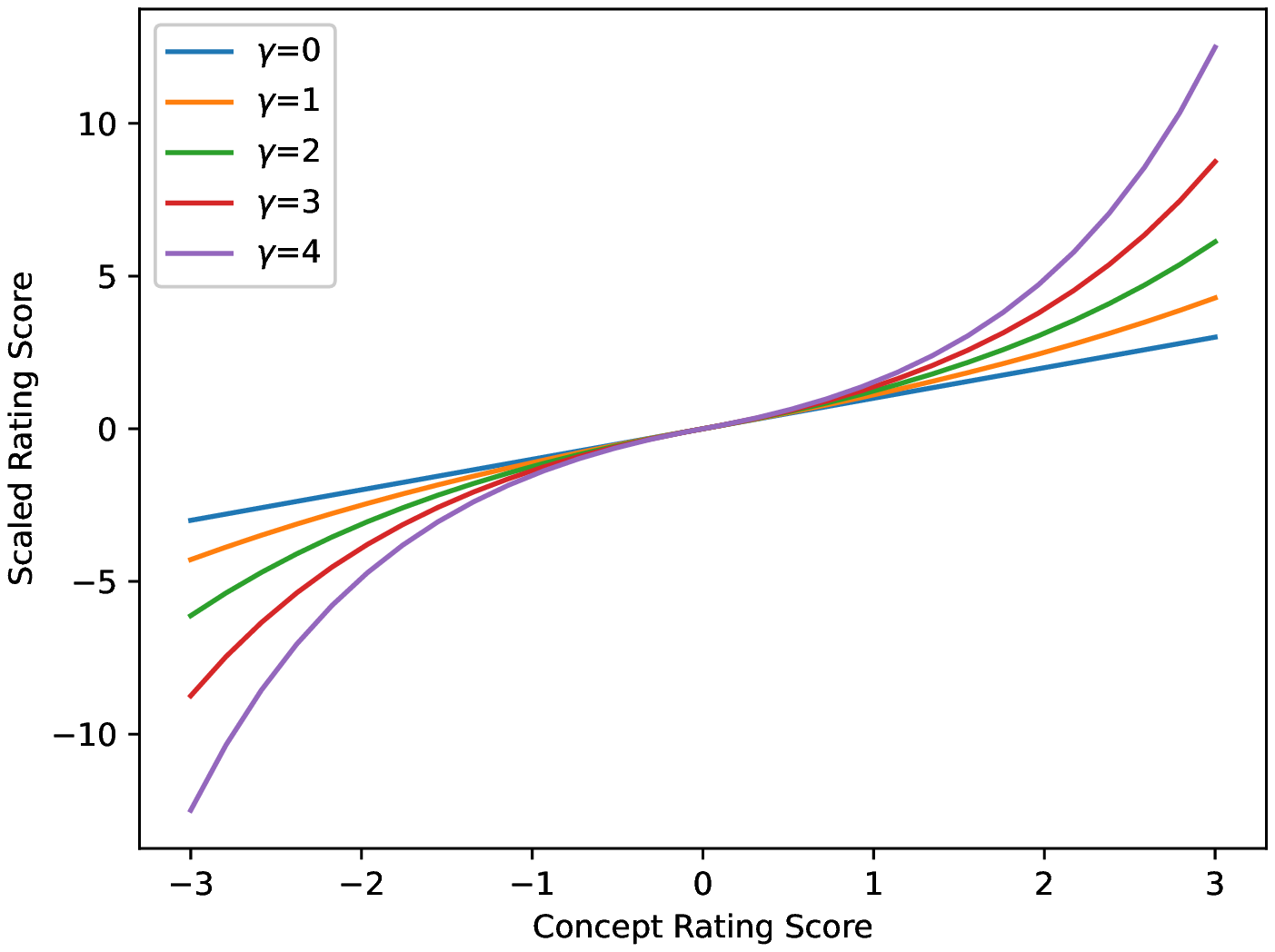}
\caption{Focal Rating impact visualization. Raw rating scores (blue line) are scaled up with amplitude proportional to the different focal factor $\gamma$ values (K fixed to 10). The attention is put more on infrequent highly rated concepts in both sides, while no significant changes are found  on dominant low rated concepts.}
\label{fig:FR}
\end{center}
\end{figure}

\section{Effect of focal rating}
Visual concept ratings were gathered using a linear scale which ranges from -3 to +3. 
The distribution of rating patterns from Figure 1 of the main paper shows that a large number of visual concepts have ratings which are close to neutral.
Moreover, a large part of nearly neutral concepts appear frequently in user photos as illustrated by the patterns which range in the middle of Table~\ref{tab:patterns}.
A direct usage of visual concept rating might lead to highly-rated but less frequent concepts being overwhelmed by nearly-neutral but frequent concept.
This happens because of all methods proposed in the paper exploit a form of averaging to derive a global user profile rating. 
Focal rating is introduced to boost the influence of highly-rated visual concepts. 
We illustrate its effect in Figure~\ref{fig:FR} with different values of $\gamma$ which are then tested during $LERVUP$ optimization.
The higher the value of $\gamma$ is, the stronger the effect of focal rating will be.


\section{Optimization of $LERVUP$ training}
The $LERVUP$ training is done separately for each situation. 
A grid search for optimal parameters is implemented to find the best configurations of the learned models.
These include the parameter search for focal rating, the random forest regression method, and a process which excludes outliers in the training data.
The value ranges of the parameters are provided in Table~\ref{tab:param}. 

Raw ratings for user profiles obtained through crowdsourcing are generally coherent but the contain outliers which might have a negative effect on the learned models. 
To remove outliers, we first mapped each user descriptor of 16-dimension vector ($\text{4 clusters}\times\text{(3 image-level attributes + 1 cluster's variance)}$) into a new feature space of two dimensions by performing PCA. 
Second, we computed the summed Euclidean distance of each point to the others and kept only the points within a radius $\epsilon$.
Finally, we ranked the points based on density of neighbors within $\epsilon$ distance from the target point and kept the G\% of user profiles which have the densest representation. 

In each fine-tuning process, we generate candidate regression models by varying values of random-forest's attributes such as bootstrap, tree depth, number of estimators, etc.  
We evaluated each candidate model precision by the cross-validation technique with 5-folds on the selected user profiles. 
The best candidate is saved at the end of fine-tuning process. 
The candidates from all process are verified afterward on a validation set to choose the most appropriate statistical model for each situation. 

\begin{table}[ht]
\begin{center}
\begin{tabular}{ |c|c|c|}
\hline
Component & Parameter & Values \\
\hline
{\multirow{2}{*}{Focal rating}} & K & 10, 15, 20, 25 \\
\cline{2-3}
& $\gamma$ & 0, 1, 2, 3, 4 \\
\hline
{\multirow{2}{*}{Outlier removal}} & $\epsilon$ & 0.05, 0.1, 0.15, 0.2 \\
\cline{2-3}
& G & 80, 85, 90, 95, 100 \\
\hline
\end{tabular}
\caption{Parameter fine-tuning values}\vspace{1ex}
\label{tab:param}
\end{center}
\end{table}

\section{Optimization of detection models}
Two neural networks are trained using the Tensorflow Object Detection API v1 on a Tesla v100 GPU which were originally designed for edge computing and for classical GPU hardware respectively.
In order to accelerate the training process and to reduce imbalance between visual object representations, we limit the number of images to 1000 per class.
We also transfer parameters from pretrained models to train the final models faster.
Both networks are trained with random horizontal flips as data augmentation technique.

The larger network is a Faster-RCNN Inception Resnet v2~\cite{ren2015faster}, atrous version, pretrained on the COCO dataset. 
We use a momentum optimizer with a manual learning rate schedule starting at 0.006 for 300000 steps, then 0.0006 for 300000 more and finally 0.00006 for the remainder of the training. 

The small network is a MobileNet v2~\cite{sandler2018}, also pretrained on the COCO dataset. 
The learning rate schedule is 0.005 for the initial learning rate, 0.0005 at step 100000 and 0.00005 at 200000. We use 8-Bit quantization aware training applied starting from step 100000, in order to speed up inference on mobile.

{\small
\bibliographystyle{ieee_fullname}
\bibliography{supplement}
}